\PassOptionsToPackage{prologue,dvipsnames}{xcolor}
\documentclass[sigconf]{acmart}
\usepackage{newtxtext,newtxmath}
\usepackage{amsmath}

\usepackage{mathtools}
\usepackage{makecell}
\usepackage{pgfplots}
\pgfplotsset{width=8cm,compat=1.9}
\pgfdeclarelayer{background}% determine background layer
\pgfsetlayers{background,main}% order of layers

\usepackage{pgfplotstable}
\usepackage[dvipsnames]{xcolor}
\usepackage[]{geometry}
\usepackage[]{textcomp}
\usepackage{mathtools}
\usepackage{xspace}
\usepgfplotslibrary{fillbetween}
\pgfplotsset{compat=1.3}
\usetikzlibrary{patterns}
\usepackage{placeins}
\usepackage{listings}
\usepackage[most]{tcolorbox}

\usepackage{filecontents}
\usepackage{booktabs}
\usepackage{color, colortbl}
\usepackage{subcaption}
\usepackage{ulem} % for strikethrough
\usepackage{multicol}
\usepackage{multirow}
\usepackage{adjustbox}

\usepackage{booktabs}

\newcommand{\lFig}[1]{\label{fig:#1}}
\newcommand{\rFig}[1]{Figure \ref{fig:#1}}
\newcommand{\lSec}[1]{\label{sec:#1}}
\newcommand{\rSec}[1]{Section \ref{sec:#1}}
\newcommand{\rTheory}[1]{\ref{sec:#1}}

\newcommand{\lTable}[1]{\label{tab:#1}}
\newcommand{\rTable}[1]{Table \ref{tab:#1}}

\newcommand{\streamingrag}{StreamingRAG\xspace}

\AtBeginDocument{%
  \providecommand\BibTeX{{%
    \normalfont B\kern-0.5em{\scshape i\kern-0.25em b}\kern-0.8em\TeX}}}

\setcopyright{acmlicensed}
\begin{document}

\acmYear{2024}\copyrightyear{2024}
\acmConference[AI4Sys '24]{Workshop on AI For Systems}{June 3--7, 2024}{Pisa, Italy}
\acmBooktitle{Workshop on AI For Systems (AI4Sys '24), June 3--7, 2024, Pisa, Italy}
\acmDOI{10.1145/3660605.3660943}
\acmISBN{979-8-4007-0652-3/24/06}

\title{\streamingrag: Real-time Contextual Retrieval and Generation Framework}

\author{Murugan Sankaradas}
\email{murugs@nec-labs.com}
\affiliation{%
  \institution{NEC Laboratories America}
  \streetaddress{4 Independence Way}
  \city{Princeton}
  \state{NJ}
  \country{USA}
  \postcode{08540}
}

\author{Ravi K. Rajendran}
\email{rarajendran@nec-labs.com}
\affiliation{%
  \institution{NEC Laboratories America}
  \streetaddress{4 Independence Way}
  \city{Princeton}
  \state{NJ}
  \country{USA}
  \postcode{08540}
}

\author{Srimat T. Chakradhar}
\email{chak@nec-labs.com}
\affiliation{%
  \institution{NEC Laboratories America}
  \streetaddress{4 Independence Way}
  \city{Princeton}
  \state{NJ}
  \country{USA}
  \postcode{08540}
}

\renewcommand{\shortauthors}{Sankaradas, et al.}

\begin{abstract}
    % abstract
Extracting real-time insights from multi-modal data streams from various domains such as healthcare, intelligent transportation, and satellite remote sensing remains a challenge. High computational demands and limited knowledge scope restrict the applicability of Multi-Modal Large Language Models (MM-LLMs) on these data streams. Traditional Retrieval-Augmented Generation (RAG) systems address knowledge limitations of these models, but suffer from slow preprocessing, making them unsuitable for real-time analysis. We propose StreamingRAG, a novel RAG framework designed for streaming data. StreamingRAG constructs evolving knowledge graphs capturing scene-object-entity relationships in real-time. The knowledge graph achieves temporal-aware scene representations using MM-LLMs and enables timely responses for specific events or user queries. StreamingRAG addresses limitations in existing methods, achieving significant improvements in real-time analysis (5-6x faster throughput), contextual accuracy (through a temporal knowledge graph), and reduced resource consumption (using lightweight models by 2-3x).

\end{abstract}

%%
%% The code below is generated by the tool at http://dl.acm.org/ccs.cfm.
%% Please copy and paste the code instead of the example below.
%%



% \received{20 February 2007}
% \received[revised]{12 March 2009}
% \received[accepted]{5 June 2009}

\maketitle

% introduction
\section{Introduction}
\lSec{introduction}
The ever-growing volume of streaming data presents great opportunities across diverse fields like healthcare for real-time analysis of medical images \cite{healthcare_vlm-hartsock-2024}; intelligent transportation systems for understanding traffic flow patterns \cite{its_survery-veres-2020}; and remote sensing, where satellite constellations provide stream of imagery for applications like, environmental monitoring, economic activity measurement and disaster response \cite{cloud_detection-czerkawski-2023, satin-roberts-2023}. These domains share a common fundamental underlying need: the ability to extract real-time insights from video, image, and other multi-modal sensor streams. Dynamic situations and anomalies are frequent occurrences in these domains which need immediate attention. MM-LLMs \cite{clip-radford-2021, blip2-Junnan-2023, show_tell-oriol-2015} are often utilized due to their ability to efficiently extract information from multi-modal streams.
Traditionally, the RAG framework \cite{rag-lewis-2021} has been used to understand static data by gathering knowledge about the situation, which relies on fetching, indexing, and converting external data into structured formats - a time-consuming process. Hence it is unsuitable for real-time applications, leading to a potential gap in understanding critical events within the data stream. Extracting information using MM-LLMs requires significant amount of resources, which increases costs. For instance, MM-LLM can take up to 3-5 seconds per video frame in case of GPT-4V \cite{gpt4v_tr-openai-2023}, which is impractical for real-time streaming applications as it can miss crucial unfolding events.

To address limitations of existing methods and achieve real-time comprehension, we propose StreamingRAG, a framework that utilizes efficient models to construct an evolving knowledge graph \cite{knowledge_graph-hogan-2021} about the stream content. StreamingRAG accomplishes this by extracting contextual scene-object-entity relationships. This knowledge graph serves two key purposes: (1) providing an efficient scene representation and (2) facilitates real-time and contextual information retrieval. Extracting all actors and their relationships in a real-time scene is computationally expensive.

Therefore, our research proposes a dynamic priority-based approach for knowledge extraction. We prioritize information about specific actors and relationships based on user constraints, evolving events, and current context. The scheduler orchestrates this dynamic prioritization and utilizes a multi-modal question generator to ask relevant questions at a given time. This approach ensures that key information about  prioritized actors and relationships is extracted within the allotted time frame, ultimately leading to the construction of knowledge graph in real-time that reflects the most crucial aspects. We make the following key contributions.
\begin{itemize}
    \item StreamingRAG, a Retrieval-Augmented Generation (RAG) system for streaming data, tackles the challenge of querying real-time data streams, by constructing scene-aware spatio-temporal knowledge graphs using lightweight models. 
    \item Context-driven dynamic priority-based knowledge extraction process, which enables the efficient construction of knowledge graph.
    \item We evaluate the effectiveness of the StreamingRAG in monitoring video streams for Intelligent Transportation System. It yields significant advantages: 5-6x faster processing (throughput) compared to prior work, ensuring real-time analysis. It maintains contextual accuracy in evolving scenarios using temporal knowledge graph. Finally, the use of lightweight models reduces resource utilization by 2-3x.
\end{itemize}

% motivation
\section{Motivation}
\lSec{motivation}
Traditional RAG systems enhance the capabilities of MM-LLMs by supplementing their knowledge with retrieved external data sources. Using semantic search, retriever module builds enriched context by fetching relevant information from external sources, which is then incorporated to enrich query context so that LLM can answer questions, create descriptions, and complete tasks by combining the multi-modal input and retrieved relevant data. Existing systems rely on computationally expensive MM-LLMs to process data and store the extracted information. The effectiveness of this approach is measured by three factors: information loss due to processing time, response delay, and computational resources required.

\begin{figure}[h!]
    \centering
    \includegraphics[width=0.95\linewidth, scale=1]{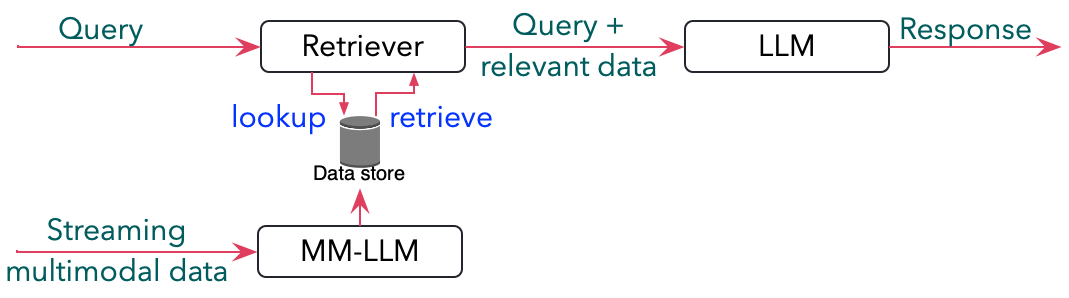}
    \caption{Baseline architecture}
    \lFig{traditional-rag-baseline}
\end{figure}

\subsection{Limitations of baseline architecture}
Heavyweight MM-LLMs have lengthy processing times, which limits their use to batch processing systems, where offline processing is done. This approach is not suited for realtime streaming scenarios where immediate analysis is needed on streaming data as it arrives. Key metrics of streaming applications are as follows.

\textit{Accuracy:} Baseline fails to generate accurate responses for an unfolding event, missing crucial details and failing to adapt evolving context. %due to stale retrieval point. 
Inaccurate timing and sequencing caused by missing to process enough data chunks, disrupt the order of events, creating time gaps and potentially misinterpreting the flow of the stream. Anomaly detection becomes challenging as crucial deviations from the norm might be masked by missing data. %, leading to missed signals or false positives.

\definecolor{greenShade}{RGB}{197, 224, 180}
\definecolor{redShade}{RGB}{248, 203, 173}
\begin{figure}[h!]
    \centering
    \includegraphics[width=0.75\linewidth, scale=1]{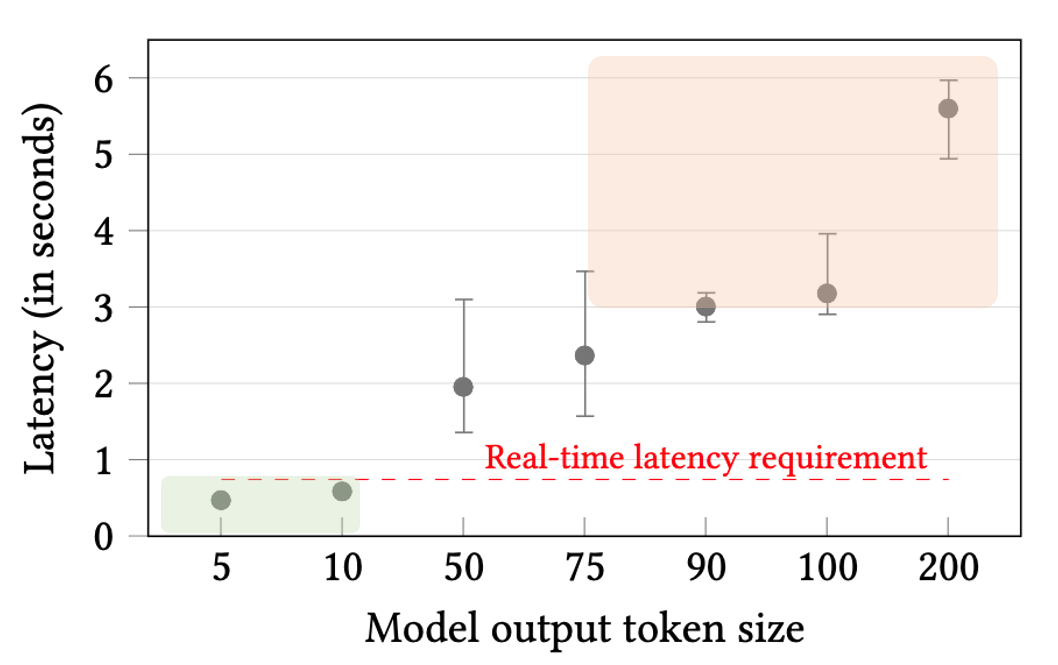}
    \caption{Latency vs output token size of ShareGPT4V \cite{sharegpt4v-chen-2023}. \textcolor{redShade}{Red} shaded area indicates descriptive answers, using larger output token size which incrases latency. \textcolor{greenShade}{Green} shared area generates less descriptive output meeting real-time constraints using smaller output token size. Need to ask descriptive questions to model so that spatio-temporal information across frames is extracted.}
    \lFig{tokens-vs-latency}
\end{figure}

\textit{Latency \& Throughput:} Powerful heavyweight models demand significant resources, leading to large execution times, leading to reduced throughput and responsiveness. Furthermore, obtaining spatial details necessitate generative models to produce more elaborate responses, potentially exceeding over 50 tokens. For instance, for a vision-to-language model, the latency associated with various output response token sizes of the ShareGPT4V VLM \cite{sharegpt4v-chen-2023} is shown in \rFig{tokens-vs-latency}. Evidently, to function in real-time, the output token sizes must be minimized, reinforcing the importance of employing tailored lightweight models to address targeted questions pertaining to descriptive inquiries.

\textit{Cost:} MM-LLMs have immense computational demands which require state of the art hardware accelerators. Processing hundreds of sensor feeds concurrently for deployments like smart city applications in real-time can quickly overwhelm available hardware. Increased latency hinders the ability of the system to provide timely insights, while reduced contextual accuracy compromises the quality and usefulness of generated responses. 

\subsection{Our approach}
Instead of using heavy duty models to extract information from real-time streaming content, \streamingrag tackles the challenge, as shown in \rFig{streaming-rag-ours}. Unlike traditional methods burdened by heavyweight models, \streamingrag prioritizes both temporal context and efficiency by systematic extraction and dynamic knowledge pipelines, as explained in \rSec{streaming-rag}. This is achieved through extracting low-level object information and their relationships instead of the high-level scene by dynamically prioritizing information about specific entity-attribute-relationship and creating embeddings for incoming streams followed by the construction of knowledge graphs. \streamingrag maintains contextual accuracy in evolving scenarios by utilizing the temporal knowledge graph. By prioritizing efficiency and context-awareness, \streamingrag enables efficient exploration of real-time data streams.
\begin{figure}[h!]
    \centering
    \includegraphics[width=0.95\linewidth, scale=1]{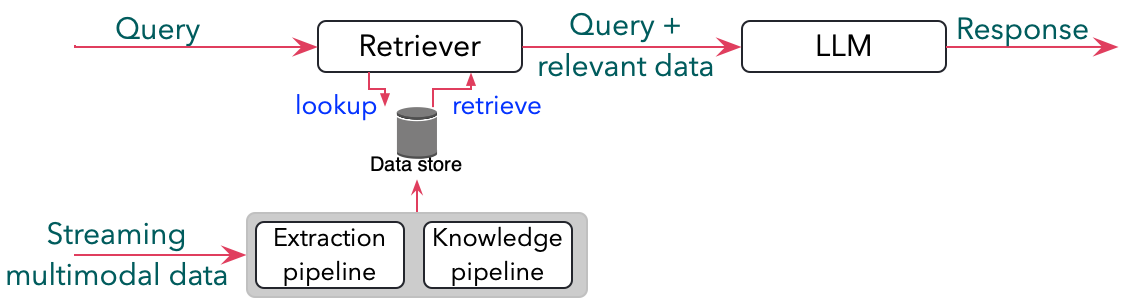}
    \caption{StreamingRAG approach}
    \lFig{streaming-rag-ours}
\end{figure}

\begin{figure*}[t]
    \centering
    \includegraphics[width=0.9\textwidth]{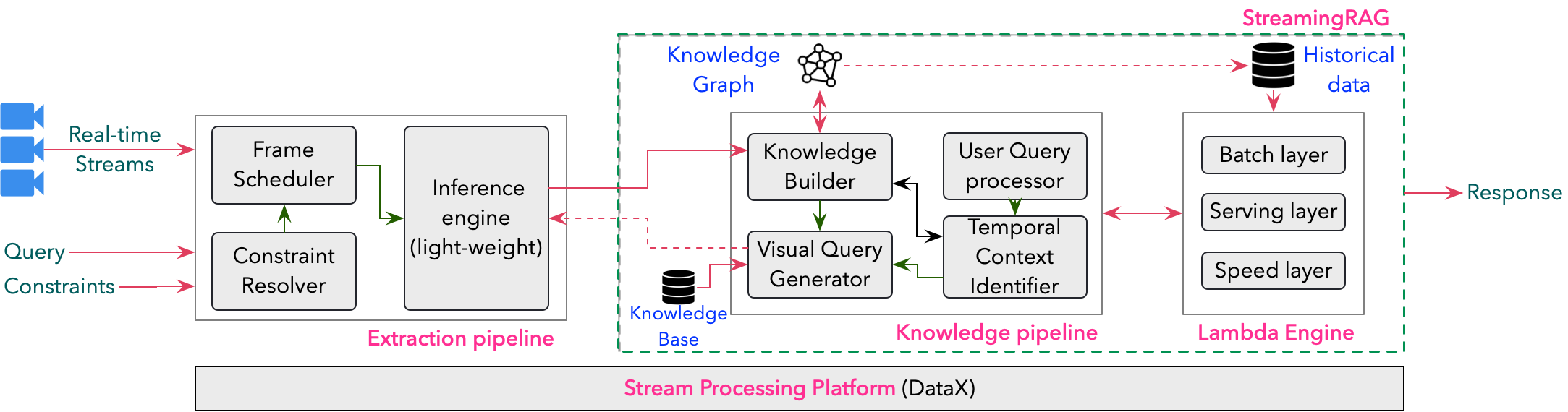}
    \caption{\streamingrag: System Architecture}
    \lFig{system-architecture}
\end{figure*}

% proposed-work
\section{Streaming RAG}
\lSec{streaming-rag}
\subsection{Overview}
\lSec{system-overview}
StreamingRAG framework handles both real-time continuous monitoring through standing queries and interactive queries. Standing queries (aka persistent queries), are ongoing queries that continuously scan real-time streams for specific updates/conditions. They are different from one-time queries with a finite response, offering a continual awareness of specific events or patterns within the data. Interactive queries, which are submitted by users on demand, require both real-time and historical data. They allow users to refine their query based on the initial response, then dive deeper into specific aspects. StreamingRAG (\rFig{system-architecture}) consists of 4 main components: (a) extraction pipeline, (b) knowledge pipeline, (c) stream processing platform, and (d) lambda engine as presented below.

\subsection{Extraction Pipeline}
\lSec{extraction-pipeline}
Spatial metadata is extracted using various inference engines (i.e., VLMs, LLMs, and MM-LLMs etc.) from frames, subject to real-time constraints. Data selection from input streams is prioritized dynamically based on evolving events, due to impracticality of processing every incoming data chunk.

\subsubsection{Frame Scheduler:}
\lSec{component-frame-scheduler}
Frames are received from streaming video sources to input stage of the extraction pipeline, which is running on top of the stream processing platform (\rTheory{component-stream-processing-platform}). These frames are then directed to models of different capabilities (\rTheory{component-inference-engine}). The scheduler analyzes incoming frames to assess their content and complexity, considering factors like motion and scene details. System-level constraints are simultaneously monitored by a Constraint Resolver (\rTheory{component-constraint-resolver}). This combined analysis allows the frame scheduler, with both prioritized rules and adaptive algorithms, to select the optimal frame rate of input data being fed to the inference engines.

\subsubsection{Constraint Resolver:}
\lSec{component-constraint-resolver}
The constraint resolver optimizes the dynamic scheduling of frames across the models within the real-time extraction pipeline to enhance system responsiveness. Factors like user-specified constraints (frames per second, model processing latency, % a predefined maximum latency threshold, 
inference cost) and system-level operational constraints (CPU usage, GPU load, memory availability) are optimized to maximize throughput while ensuring latency constraints. 

\subsubsection{Inference Engine:}
\lSec{component-inference-engine}
The real-time extraction pipeline uses a combination of heavyweight and lightweight Multi-Modal Large Language Models. The models take in frames and list of queries to be processed on the given frame. The VLMs are specifically dedicated to handling questions from the question bank for a tiered analysis, acting as a filter for any event detection by optimizing computational resources.

\subsection{Knowledge Pipeline}
While extraction pipeline extracts only the metadata from input streams, knowledge pipeline translates the metadata into actionable knowledge, enabling the system to answer user queries and, more importantly, provide feedback to the extraction pipeline, guiding its metadata extraction from incoming streams. These include analyzing the current response, constructing temporal context using spatial details across frames, monitoring events as they unfold, refining queries for subsequent frames. The functional components in knowledge pipeline are described below.

\subsubsection{Knowledge Base:} 
Knowledge Base ($KB$) serves a foundational repository of common-sense knowledge, providing an initial context for understanding incoming streaming data. It aids in Knowledge Graph initialization \cite{google_kg_search_api-google-2022, amazon_knowledge_graphs-amazon-2021, dbpedia-Auer-2007, geonames-Geonames-2005}, forming a starting point for dynamic context building. $KB$ depends on the use cases. For instance, in the traffic scenario, $KB$ contains information such as different actors like pedestrians, people, drivers, objects like vehicles and their relationships, traffic rules, and historical traffic patterns. 

\subsubsection{Knowledge Graph:}
A knowledge graph ($KG$) is represented using semantic tuple of three elements: subject, predicate, and object. The subject and object represent an entity pair, such as a person-location, or an object-property. The predicate specifies the relationship connecting these entities. $KG$ acts as the dynamic memory, capturing the contextual details necessary for response generation. The information within the $KG$ is contingent upon the context of the specific use case, and it can be sourced from the $KB$. 

\subsubsection{Knowledge Builder:} 
Knowledge Builder updates and refines the $KG$ in real-time based on $KB$ and user-directed contexts. It assembles entities, relationships, and contextually relevant details from the $KB$ into the evolving $KG$. It continuously updates the graph, reflecting evolving context, and adapts to influx of streaming data, ensuring that the graph remains reflective of the real-time event, allowing downstream components, such as the retriever and language models, to operate on up-to-date context. Generating $KG$ from the scene-level spatial understanding is achieved by modeling $Pr(G | F)\ =\ Pr(B, L, R)$, where $F$ is the input frame, $G$ is the desired graph, $b_i \in B$ is the bounding box in the images, $L$ is the object labels and $R$ is the relations among the objects $L$.

\subsubsection{Temporal Context Identifier:}
\lSec{component-context-identifier}
Understanding incoming streams requires establishing temporal context (e.g., event or incident). This context acts as a trigger to schedule resource allocation, such as processing power (CPU/GPU) or higher frame rates, ensuring critical details aren't missed during crucial moments. Combining spatial information extracted from each frame in the extraction pipeline provides a comprehensive understanding of the temporal context. Extracting temporal context involves iteratively following these steps: Identify the user query's specific needs; Fetch relevant records from the data store; Prioritize retrieved content through ranking and filtering via moderation. Identifying context and extracting necessary information from incoming frames involves querying the knowledge base (KB) to identify associated entities and relationships. Entities with the highest probabilities are selected, prompting the construction of prompts for subsequent iteration of questions within the VLM inference engine in the extraction pipeline. Once an event concludes or under steady-state conditions, the KG undergoes reset, ensuring alignment with the current state and to prepare for subsequent events and queries.

\subsubsection{Visual Query Generator:}
\lSec{component-visual-query-generator}
$VQG$ is an interface between extraction and knowledge pipelines, incorporating the current video response to pull pertinent information from $KG$. Given the current video response $S_{t}$ and questions $Q_{t}$, it interfaces with $KG$ through the temporal context identifier at the time instant $t$. Query generator integrates $S_t$ with the $KB^t$ to update the set of questions, $Q_{t+k} \leftarrow VQG(S_t, KB^t)$ where $k > 0$ based on the temporal cues.
This process refines the queries, incorporating the evolving context. The refined questions, $q \in Q_{t+k}$, are then scheduled for the subsequent frames $f_{t+k}$ are then embedded into the current prompt $P$, which is dispatched to the extraction pipeline using the scheduler. %The process is streamlined to ensure the continuous adaptation of queries to the evolving event-based context.

\subsubsection{User Query Processor:}
\lSec{component-user-query-processor}
Query processor translates the unstructured, multi-modal user queries into actionable executors for the pipeline.
In addition to parsing and interpreting user queries, processor refines queries based on the evolving context and $KB$. 

\subsection{Stream Processing Platform}
\lSec{component-stream-processing-platform}
The system components are hosted on real-time stream processing platform DataX \cite{datax-coviello-2022}. It tackles the complexity of building distributed stream processing applications by streamlining data exchange, transformations, and fusion. DataX provides an abstraction, which simplifies application specification by exposing parallelism and dependencies among microservices. This allows the DataX runtime to automatically manage data communication and provides dynamic scaling and optimization.

\subsection{Lambda Engine}
\lSec{component-lambda-engine}
Lambda engine is used to support both real-time standing queries and interactive queries. The framework requires access to both real-time and historical data, and this access is expedited to minimize query response time through three layers: batch, serving and speed layer. The batch layer pre-processes and updates $KG$ with historical context, providing foundation for contextual understanding. This result is then infused to the real-time extraction and knowledge pipelines. The serving layer enables efficient retrieval of relevant data from $KG$, enhancing the retrieval process for user's interactive queries. The speed layer handles real-time data streams, so that system is adaptive and responsive to evolving contexts.
% experimental-evaluation
\section{Experiments \& Evaluations}
\lSec{experiments-and-evaluation}
\subsection{Setup}
\lSec{eval-setup}
StreamingRAG is evaluated in the context of Intelligent Transportation Systems (ITS) for anomaly event detection by generating event descriptions, using both public \cite{anomaly_dataset-paperswithcode} and proprietary dataset, for various event complexities, with a specific focus on evaluating real-time evolving events through following standing queries. \textit{(1) Raise an alert when there is an accident involving collision between vehicles and people. (2) Inform me upon spotting pedestrians in distress.}

Our experimental setup consists of a cluster of servers, each equipped with a 32-core Intel Xeon CPU, 256 GB of memory, and 24 GB NVIDIA GeForce RTX 3090 Ti. Our evaluation methodology involves streaming the videos captured by an IP camera. 

\textit{Extraction pipeline} utilizes ShareGPTV4 \cite{sharegpt4v-chen-2023}, a descriptive image captioning model, for inference in our baseline system. VQA models, BLIP \cite{blip-Junnan-2022} and BLIP-2\cite {blip2-Junnan-2023}, are used to extract spatial information.

\textit{Streaming Processing Platform} is built using DataX \cite{datax-coviello-2022} real-time stream processing framework. It enables us to efficiently manage the continuous data flow from camera feeds, process it using the chosen VLM models, and integrate it with knowledge pipeline. 

Inside \textit{Knowledge Pipeline}, $KB$ is manually created for our traffic monitoring use case, currently implemented as a mapping between entities, potential attributes and relationships among them. $KG$ is maintained using the NetworkX \cite{networkx-hagberg-2008}. Upon receiving responses from the VLM for the previous set of VQG generated queries, the generator retrieves relevant entities, attributes and necessary relationships between entities, from $KB$. Subsequently, it formulates a prompt comprising a set  of upto five questions, depending on the scenario context, to stay within the real-time latency requirement. 

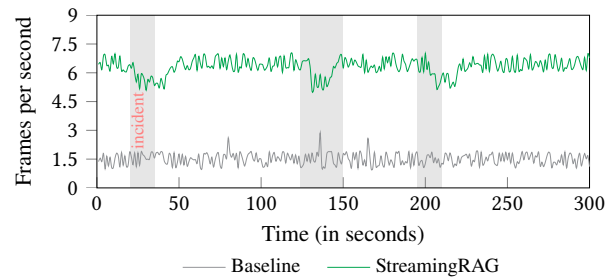
\begin{figure}[h!]
    \centering
    \begin{tikzpicture}[transform shape, scale=0.95]
        \begin{axis}[
            width=1\linewidth,
            height=4cm,
            xlabel={Time (in seconds)},
            ylabel={Frames per second},
            xmin=0, xmax=300,
            ymin=0, ymax=9,
            xtick pos=bottom, ytick pos=left,
            ytick align=outside,
            ytick={0,1.5,...,9},
            legend style={draw=none, nodes={font=\small, scale=1, transform shape}},
            legend style={at={(0.5,-0.35)},anchor=north},
            legend columns=-1,
            legend style={/tikz/every even column/.append style={column sep=0.3cm}},
        ]
        \begin{pgfonlayer}{background}
            \fill[color=black!10] (axis cs:20,0) rectangle (axis cs:35,9);
            \fill[color=black!10] (axis cs:124,0) rectangle (axis cs:150,9);
            \fill[color=black!10] (axis cs:195,0) rectangle (axis cs:210,9);
        \end{pgfonlayer}
        \node[draw=none, align=center, xshift=-0.2cm, color=red!50, font={\footnotesize}, rotate=90, anchor=north] 
                at (axis cs:25, 3.5) {incident};
        \addplot[smooth,Gray] table [x=idx, y=baseline, col sep=comma]{data/fps_over_time.txt};
        \addlegendentry{Baseline};
        
        \addplot[smooth,Green] table [x=idx, y=streamingrag, col sep=comma]{data/fps_over_time.txt};
        \addlegendentry{StreamingRAG};
        \end{axis}
    \end{tikzpicture}
    \caption{Throughput}
    \lFig{fps-vs-time}
\end{figure}

\subsection{Evaluation Metrics}
\lSec{eval-metrics}
Our evaluation considers both the duration of the event and its complexity. Duration of an event is critical for the real-time analysis because if a frame takes a longer time to process, a short-time event will be completely missed. Long-time events are more manageable in the real-time extraction phase, due to system's ability to observe and formulate questions for subsequent frames. System performance is evaluated using the following metrics.

\textit{Latency \& Throughput} -- Time taken to generate responses. %is one of the two system evaluation metrics
% on the efficiency of StreamingRAG. Heavyweight VLMs are not ideal for the real-time applications, acting as a
% which is a bottleneck that limits overall system performance.

\textit{Cost (GPU resource utilization)} is measured in terms of GPU memory usage, a factor determining the scalability of the system.

\textit{Accuracy} of generated responses is evaluated through human examination, with a specific focus on system's adeptness at identifying evolving contexts and creating responses. In our use case, evaluation is done on two factors (a) context-aware event detection (b) response generation. For instance, in hit-and-run, the system needs to identify the exact sequence: vehicle-person collision, immobility of person following the collision and vehicle fleeing the scene. Missing any event loses the context, leading to an incorrect response. Generation phase assesses the accuracy and adequacy of model's responses (both LLMs and VLMs). 

\subsection{Results}
\lSec{eval-results}
\subsubsection{Latency \& Throughput:} The shaded areas in the graph (\rFig{fps-vs-time}) denote instances of anomaly incidents such as hit-and-run and vehicle collisions. 
In baseline, higher inference times causes crucial temporal information loss when frame-level responses are aggregated to create response. However, StreamingRAG operates at approximately 8 fps ($1/3^{rd}$ of video's fps), effectively leading to extract contextual spatial information. 

\subsubsection{GPU resource utilization:} While \rTable{gpu-resource-requirement} highlights the efficiency gains of StreamingRAG for a single camera stream with a dedicated GPU, real-world deployments involve hundreds of cameras distributed across a wide area. Our architecture allows GPU sharing, to dynamically adjust resource allocation based on the evolving context of each stream, ensuring efficient resource utilization even in complex and geographically distributed deployments. 

\begin{table}[h!]
    \caption{GPU resource utilization}
    \lTable{gpu-resource-requirement}
    \begin{adjustbox}{width=0.6\columnwidth,center}
    \setlength{\tabcolsep}{0.1cm}
    \begin{tabular}[t]{lccc}
        \hline
         &\makecell{Baseline} & &\makecell{With StreamingRAG}\\
         \hline
         \hline
         VLM & 18GB && 8GB\\
         LLM Summarizer & - && 2GB\\
         \hline
         Total & 18GB && 10GB\\
    \end{tabular}
    \end{adjustbox}
\end{table}

\subsubsection{Accuracy:} In real-time scenarios, the ability to promptly detect unfolding incidents is critical. We achieved a notably higher detection rate as shown in \rTable{incident-detection}, due to its faster processing capabilities and the utilization of historical frame-level responses. Despite this improvement, both systems encountered challenges in promptly detecting short-term events, although ours demonstrated a relatively higher success rate compared to baseline. Generated responses are evaluated based on accuracy of the chronological sequence of events accumulated temporally. \rTable{results-description-generation-comparison-various-scenarios} shows the responses of 4 events such as hit-and-run, vehicle-to-vehicle collision, vehicle-to-pedestrian collision and people commotion. The baseline struggles to process frames promptly, resulting in a failure to gather information about the individuals and vehicles involved, ultimately missing the collision events entirely. For instance, in Vehicle-to-Vehicle collision, the baseline correctly identified that a car was on fire. However, it failed to capture the context of the collision between vehicles, which resulted in the car catching on fire. In vehicle-to-pedestrian collisions, the baseline not only failed to detect the collision event but also misclassified the situation, interpreting it as a game rather than within the context of a traffic scenario. With StreamingRAG, in hit-and-run case, the exact moment of the accident was overlooked. However, through analyzing multiple frames, it found the person was lying on the road near the car, indicating the occurrence of an accident (highlighted in red).

\begin{table}
    \caption{Results of event detection on various event types}
    \lTable{incident-detection}
    \begin{adjustbox}{width=1\columnwidth, center}
    \begin{tabular}{lccccc}
        \toprule
        \multicolumn{4}{c}{Anomaly Event} & \multicolumn{2}{c}{Events detected}
        \\\cmidrule(lr){1-4}\cmidrule(lr){5-6}
        \makecell{Type} & \makecell{No. of Videos} & 
        \makecell{Average Duration\\(in seconds)} & 
        \makecell{Ground truth\\(number of events)} & 
        \makecell{Baseline} & \makecell{StreamingRAG}\\
       \midrule
         Hit and run                        & 2 &  8 & 3 & 0 & 2 \\
         Vehicle-to-Vehicle collision       & 2 & 16 & 3 & 1 & 3 \\
         Vehicle-to-Pedestrian collision    & 1 & 48 & 1 & 0 & 1 \\
         People commotion                   & 1 & 26 & 1 & 0 & 1 \\
        \bottomrule
            &  &  &  Total events: 8 &  1 (12.5\%) & 7 (87.5\%) \\
    \end{tabular}
    \end{adjustbox}
\end{table}

\begin{table}[h!]
    \caption{Accuracy evaluation of baseline and StreamingRAG based on context-aware response generation}
    \lTable{results-description-generation-comparison-various-scenarios}
    \begin{adjustbox}{width=\linewidth, center}
    \tiny
    \begin{tabular}{lp{0.38\linewidth}p{0.38\linewidth}}
        \toprule
        \makecell{Event} & \multicolumn{2}{c}{Generated responses}
        \\\cmidrule(lr){1-1}\cmidrule(lr){2-3}
        Type (Duration) & \makecell{Baseline} & \makecell{StreamingRAG}\\
       \midrule
         \makecell[l]{Hit and run (8s)} & The image captures a winter scene on a city street. Two individuals are seen walking on the sidewalk, bundled up against the cold. A car is parked on the street, its surface dusted with snow. The street itself is lined with trees, their branches bare, a common sight in winter. The sky overhead is a blanket of gray, hinting at the possibility of more snowfall. The overall atmosphere is quiet and serene, a typical winter's day in a city. & The scene has good daylight and the background consists of a woman in a red coat and red hat standing on the corner of a city street. There are trucks present as vehicles. In the image, a woman in red pants and a red jacket is seen standing on the street corner. \textcolor{red!50}{\textit{There is a person lying down on the street near a car, indicating a potential accident}} \\
         \makecell[l]{Vehicle-to-Vehicle\\collision (11s)} & \textcolor{red!50}{\textit{A car is on fire}} & The scene depicts a highway at night. People in the scene are observed driving on the highway, where cars and trucks are present. Notably, \textcolor{red!50}{\textit{two cars are adjacent to each other, with one of them engulfed in flames within the image, suggesting a potential collision between the vehicles.}} \\
         \makecell[l]{Vehicle-to-Pedestrian\\collision (48s)} & The image captures a scene of urban decay, dominated by a pile of rubble and debris. The rubble, a chaotic mix of bricks, concrete, and other construction materials, is scattered haphazardly across the street. Amidst this disarray, a yellow sign stands out, bearing the words "Games for Windows" and "Games for Windows Live". The sign, once a beacon of digital entertainment, now seems out of place amidst the physical detritus. % In the background, a building with a graffiti-covered wall adds to the overall sense of decay. %The image is taken from a low angle, looking up at the rubble and the sign, emphasizing the scale of the destruction and the stark contrast between the digital and the physical worlds. 
         & The scene depicts a dark daylight setting with a background consisting of a pile of garbage. Various types of vehicles, particularly those used for food transportation, are present in the area. \textcolor{red!50}{\textit{There may be a potential crash because the image shows a chaotic scene with rubble scattered around and a car crushed by the debris. People in the scene are observed attempting to save their lives amidst the chaotic environment.}} \\
         \makecell[l]{People commotion\\(26s)} & The image captures a scene in a parking lot. A police officer, dressed in a black uniform, stands in the center of the frame, holding a walkie-talkie. The officer is positioned in front of a silver SUV, which is parked in a handicapped spot. The parking lot is filled with several other vehicles, including a black van and a red truck. % In the background, a building with a green roof can be seen. The image is taken from an aerial perspective, providing a comprehensive view of the scene. 
         & The scene features a variety of vehicles present in a parking lot environment. The background consists of this parking lot, \textcolor{red!50}{\textit{where a man walking his dog was struck by a car}}. The scene is illuminated with good daylight. In the image, a car is depicted within the parking lot alongside individuals on the ground and \textcolor{red!50}{\textit{some engaged in a physical altercation. Additionally, police presence is noted within the scene.}} \\
        \bottomrule
    \end{tabular}
    \end{adjustbox}
\end{table}

% conclusion
\section{Conclusion}
\lSec{conclusion}
In this paper, we addressed the limitations of high processing demands and latency associated with using Multi-Modal Large Language Models for real-time information extraction from streaming data. We proposed \streamingrag, a framework that leverages lightweight models to construct an evolving knowledge graph capturing scene-object-entity relationships. Our results demonstrate significant improvements in terms of contextual temporal accuracy, reducing the resource utilization and costs. Additionally, \streamingrag facilitates interactive querying, which enables real-time decision-making in critical scenarios.

% \newpage
%% The next two lines define the bibliography style to be used, and
%% the bibliography file.
\bibliographystyle{ACM-Reference-Format}
\bibliography{references}

\end{document}